\documentclass[letterpaper, 10 pt, conference]{ieeeconf}  

\IEEEoverridecommandlockouts                              

\usepackage{cite}
\usepackage{amsmath,amssymb,amsfonts}
\allowdisplaybreaks
\usepackage{graphicx}
\usepackage{textcomp}
\usepackage{xcolor}
\usepackage[english]{babel}
\usepackage{bm}
\usepackage{array}
\usepackage{textcomp}
\usepackage{stfloats}
\usepackage{url}
\usepackage{soul}
\usepackage{hyperref}
\usepackage{verbatim}
\usepackage{graphicx}
\usepackage{multirow}
\usepackage{algpseudocode}
\usepackage[linesnumbered, ruled, vlined]{algorithm2e}

\title{\LARGE \bf
Space-Time Graphs of Convex Sets for Multi-Robot Motion Planning
}

\author{Jingtao Tang, Zining Mao, Lufan Yang and Hang Ma
\thanks{$^{1}$The authors are with the School of Computing Science, Simon Fraser University, Burnaby, BC V5A1S6, Canada. {\tt\footnotesize \{jingtao\_tang, zining\_mao, lufan\_yang, hangma\}@sfu.ca}.}%
}

\begin{document}

\maketitle


\begin{abstract}
We address the Multi-Robot Motion Planning (MRMP) problem of computing collision-free trajectories for multiple robots in shared continuous environments. While existing frameworks effectively decompose MRMP into single-robot subproblems, spatiotemporal motion planning with dynamic obstacles remains challenging, particularly in cluttered or narrow-corridor settings. We propose Space-Time Graphs of Convex Sets (ST-GCS), a novel planner that systematically covers the collision-free space-time domain with convex sets instead of relying on random sampling. By extending Graphs of Convex Sets (GCS) into the time dimension, ST-GCS formulates time-optimal trajectories in a unified convex optimization that naturally accommodates velocity bounds and flexible arrival times. We also propose Exact Convex Decomposition (ECD) to “reserve” trajectories as spatiotemporal obstacles, maintaining a collision-free space-time graph of convex sets for subsequent planning. Integrated into two prioritized-planning frameworks, ST-GCS consistently achieves higher success rates and better solution quality than state-of-the-art sampling-based planners---often at orders-of-magnitude faster runtimes---underscoring its benefits for MRMP in challenging settings. Project page: \url{https://sites.google.com/view/stgcs}.
\end{abstract}


\section{Introduction}

We study Multi-Robot Motion Planning (MRMP), where the problem is to compute collision-free trajectories that move multiple robots from given start to goal states in space-time, while avoiding collisions both with the environment and with each other. Recent developments in Multi-Agent Path Finding (MAPF) on discrete graphs have produced powerful frameworks that decouple MRMP into single-robot trajectory computations. However, when these single-robot planners must navigate ``dynamic obstacles'' (i.e., the trajectories of other robots), the resulting spatiotemporal motion planning problem remains challenging and underexplored.

Sampling-based planners, such as Rapidly-Exploring Random Trees (RRT)~\cite{lavalle1998rapidly} and Probabilistic Roadmaps (PRM)~\cite{kavraki1996probabilistic}, are popular for their simplicity and theoretical completeness. However, spatiotemporal motion planning introduces additional challenges that significantly degrade their effectiveness: Dynamic obstacles can open or close narrow corridors, and random sampling may fail to capture these brief windows of safe transit. Moreover, many sampling-based planners either discretize time coarsely or incrementally adjust time bounds, limiting both effectiveness and efficiency in seeking time-optimal solutions within an arbitrarily large time dimension. Additionally, repeated collision checks required for state expansions become prohibitively expensive when the time dimension is included.
When applied to MRMP in a coupled manner, sampling-based planners face even greater challenges due to the curse of dimensionality, which severely restricts scalability as the concatenated state space grows rapidly with the number of robots.

In this paper, we propose Space-Time Graphs of Convex Sets (ST-GCS), a novel time-optimal motion planner that significantly improves MRMP solving. ST-GCS offers a fundamentally different, deterministic approach by systematically covering the entire collision-free space-time region with convex sets, inherently capturing spatiotemporal bottlenecks and avoiding the pitfalls of random sampling. ST-GCS extends Graphs of Convex Sets (GCS)~\cite{marcucci2024shortest}---originally designed for static, single-robot motion planning---into the spatiotemporal and multi-robot context. The key idea is to solve a generalized shortest-path problem on a graph whose vertices are convex sets, determining which sets form the path and the state within each set, which jointly optimizes a chosen objective function.

\begin{figure}
\centering
\includegraphics[width=\linewidth]{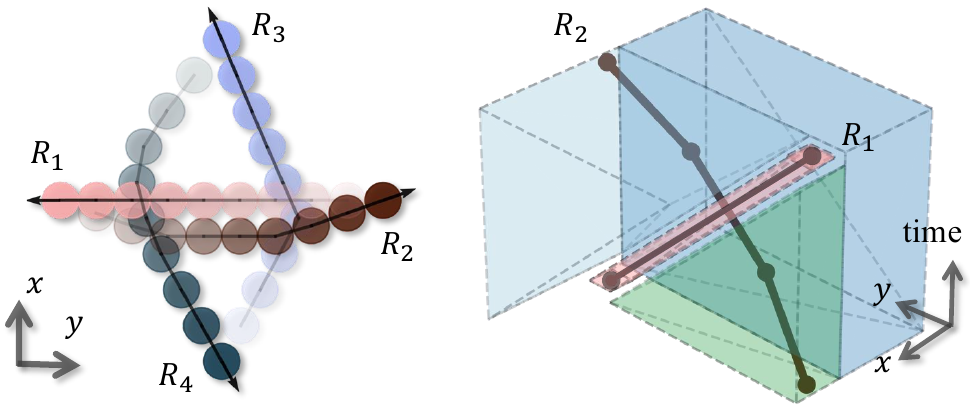}
\caption{Demonstration of the proposed PBS+ST-GCS for MRMP, where $R_1$ and $R_3$ exchange positions with $R_2$ and $R_4$, respectively. Left: Solution trajectories visualized in 2D spatial coordinates, with higher transparency indicating states at later time stamps. Right: Solution trajectories of $R_1$ and $R_2$ visualized in 3D space-time coordinates, where $R_2$ treats $R_1$ as a space-time obstacle and generates a trajectory through space-time collision-free convex sets (colored polyhedra) that exclude the trajectory of $R_1$.}
\label{fig:teaser}
\end{figure}

\noindent\textbf{Algorithmic Contributions:}
(1) We present the key idea of ST-GCS by demonstrating how to augment a spatial decomposition (i.e., a graph of convex sets) with an arbitrarily large time dimension for time-optimal trajectories. By enforcing constraints on time flow and velocity bounds within a unified convex optimization, we obtain piecewise-linear, time-optimal trajectories without specialized time discretization.
(2) We propose two MRMP methods by integrating ST-GCS into Random-Prioritized Planning (RP)~\cite{erdmann1987multiple}, which randomly explores priority orders and plans each robot sequentially according to each priority order, and Priority-Based Search (PBS)~\cite{ma2019searching}, which searches over priority orders. In both frameworks, ST-GCS serves as the low-level trajectory planner once higher-priority trajectories are determined. 
(3) To make ST-GCS tractable when higher-priority trajectories are already planned, we introduce the Exact Convex Decomposition (ECD) algorithm to partition the space-time convex sets so that subsequent planning does not conflict with these ``reserved'' trajectories. Fig.~\ref{fig:teaser} illustrates how a robot's trajectory is generated by applying ST-GCS on convex sets that have been updated to incorporate reserved trajectories as obstacle regions.

\noindent\textbf{Empirical Findings:}
We evaluate our MRMP methods against baselines that integrate a state-of-the-art sampling-based planner in a 2D mobile robot scenario. Our results demonstrate how restricting the collision-free region to space-time convex sets grants ST-GCS a unique advantage, especially in ``narrow corridors'' or crowded settings. In contrast, sampling-based methods require extensive exploration and often struggle to capture spatiotemporal safe transient windows, even when they are modified to sample directly in collision-free convex sets. Our methods consistently achieve higher success rates and better solution quality with orders of magnitude faster runtimes, highlighting the benefits of a deterministic convex-optimization approach to MRMP.

\section{Related Work}

We survey relevant research on MAPF, MRMP, and GCS.

\noindent\textbf{Search-Based MAPF:} 
MAPF \cite{stern2019multi,tan2025reevaluationlargeneighborhoodsearch} is a prominent approach for multi-robot path planning on graphs (e.g., 2D grids~\cite{solovey2016finding} or state lattices~\cite{yan2024multi}) typically assuming discrete time steps. 
Modern search-based MAPF methods have offered powerful bi-level frameworks that decompose a multi-robot planning problem into the high-level coordination (e.g., PP~\cite{erdmann1987multiple}, Conflict-Based Search~\cite{sharon2015conflict}, or PBS~\cite{ma2019searching}) that resolves collisions among individual trajectories and low-level single-robot trajectory computations (e.g., space-time A$^*$\cite{silver2005cooperative}) that respects spatiotemporal constraints posed by the high level for collision resolution. Although some MAPF methods have been adapted to continuous-time robot actions~\cite{cohen2019optimal,andreychuk2022multi,kottinger2022conflict}, their solution quality remains constrained by the chosen discrete graph representation and limited motion primitives.

\noindent\textbf{Sampling-Based MRMP:} Sampling-based MRMP methods offer greater representational flexibility by requiring only a collision checker for the environment. Coupled sampling-based MRMP approaches~\cite{solovey2016finding,solis2021representation} plan in the joint state space of all robots but rely on synchronized robot actions to facilitate collision checks; they thus do not offer time optimality in general. Several recent spatiotemporal motion planners have been combined with PP~\cite{erdmann1987multiple} for MRMP. For example, Time-Based RRT~\cite{sintov2014time} augments RRT~\cite{lavalle1998rapidly} with a time dimension but assumes a fixed arrival time at the goal state. Temporal PRM~\cite{huppi2022t} extends PRM~\cite{kavraki1996probabilistic} using safe time intervals~\cite{phillips2011sipp} but relies on constant velocity magnitude, thus compromising solution quality. 
Space-Time RRT (ST-RRT$^*$)\cite{grothe2022st} incorporates bidirectional tree search\cite{kuffner2000rrt} into RRT$^*$~\cite{karaman2011sampling} and uses a specialized conditional sampler that progressively tightens the goal arrival time bound whenever a better feasible solution is found. Although ST-RRT$^*$ is asymptotically optimal with sufficient sampling, it shares a common limitation with other sampling-based planners: ``narrow corridors'' in the space-time state space remain difficult to sample and connect. Consequently, such methods can be effective in relatively open environments but may struggle in cluttered or heavily constrained instances.

\noindent\textbf{GCS Applications:} Techniques for constructing collision-free convex sets~\cite{deits2015computing,werner2024approximating, dai2024certified} have enabled GCS-based solutions to various robotics tasks, including single-UAV path planning in cluttered environments~\cite{marcucci2024fast}, non-Euclidean motion planning on Riemannian manifolds for mobile manipulators~\cite{cohn2023non}, and temporal-logic motion planning in high-dimensional systems~\cite{kurtz2023temporal}.
Additionally, several search-based methods~\cite{chia2024gcs,natarajan2024ixg} have been developed to improve the efficiency of GCS solving.
Although \cite{marcucci2023motion} allows planning in the joint configuration space of two robotic arms with synchronous actions and \cite{von2024using} uses GCS solution to guide nonconvex trajectory optimization for dynamic environments, applying GCS to dynamic environments and multi-robot settings with asynchronous robot actions remains under-explored.

\section{Space-Time Graphs of Convex Sets (ST-GCS)}\label{sec:tgcs}
In this section, we first present the GCS formulation for single-robot motion planning around static obstacles (Sec.~\ref{subsec:gcs}). We then introduce ST-GCS, which augments GCS with an explicit time dimension for time-optimal spatiotemporal motion planning
(Sec.~\ref{subsec:gcs_time_ext}).

\subsection{GCS: Motion Planning in Time-Invariant State Spaces}\label{subsec:gcs}
We consider a single-robot motion planning problem in a $d$-dimensional state space whose collision-free region is decomposed into a given collection of convex sets (e.g., via different preprocessing techniques~\cite{deits2015computing,werner2024approximating, dai2024certified}).
This decomposition is represented by a connected graph $\mathcal{G}=(\mathcal{V},\mathcal{E})$, where each vertex $v\in\mathcal{V}$ corresponds to a convex set $\mathcal{X}_v =\{\mathbf{x}\in\mathbb{R}^d\,|\,A_v\mathbf{x}\preceq b_v\}$ and each edge $e=(u,v)\in\mathcal{E}$ indicates $\mathcal{X}_u\cap\mathcal{X}_v\neq\emptyset$.
Since $\mathcal{G}$ is connected, a simple (acyclic) \textit{path} $\pi=(v_1,v_2,...,v_{|\pi|})$ exists between any two vertices $v_1$ and $v_{|\pi|}$, using the set of edges $\mathcal{E}(\pi)=\{(v_{i-1},v_{i})\}_{i=2}^{|\pi|}\subseteq\mathcal{E}$.

Following~\cite{marcucci2023motion}, we formulate a bilinear program for the single-robot motion planning problem over $\mathcal{G}$, which is then transformed into a mixed-integer convex program for solving.
We introduce (1) a set of binary variables $\Phi=\{\phi_e\}_{e\in\mathcal{E}}$ that parameterizes any path $\pi_\Phi$, with $\phi_e=1$ if and only if $e\in\mathcal{E}(\pi_\Phi)$; and (2) two vectors of continuous variables $\mathbf{x}_{v}, \mathbf{y}_{v}\in\mathcal{X}_v$ that represent the robot's initial and terminal states within each convex set $\mathcal{X}_v$ for all $v\in \mathcal{V}$. For simplicity, we describe only the bilinear formulation below:
\begin{align}
\min_{\Phi,\mathbf{x},\mathbf{y}}\quad
& \sum_{e\in\mathcal{E}(\pi_\Phi)}f(e)+\sum_{v\in\pi_\Phi}g(v)\label{eqn:gcs:obj}\\
\textbf{s.t.}\quad
&\mathcal{E}(\pi_\Phi)\subseteq\mathcal{E},&\label{eqn:gcs:path}\\
&\mathbf{x}_v, \mathbf{y}_v\in\mathcal{X}_v,&\forall v\in\mathcal{V}\label{eqn:gcs:in_set}\\
&\mathbf{x}_v=\mathbf{y}_u,&\forall e=(u,v)\in\mathcal{E}\label{eqn:gcs:continuity}\\
&\mathbf{x}_{v_\text{start}}=\mathbf{x}_\text{start}, \mathbf{y}_{v_\text{goal}}=\mathbf{x}_\text{goal}\label{eqn:gcs:init_cond}.
\end{align}
where $f(e)$ and $g(v)$ in Eqn.~(\ref{eqn:gcs:obj}) specify additive costs over edges and vertices along the parametrized path $\pi_\Phi$.
Given a feasible solution that fixes the values of $\pi_\Phi$, $\mathbf{x}$, and $\mathbf{y}$, we can reconstruct a continuous, collision-free \textit{trajectory} $\tau$ by chaining the segments $(\mathbf{x}_v,\mathbf{y}_v)$ for each successive vertex in $\pi_\Phi$.
Specifically, Constraints Eqn.~(\ref{eqn:gcs:path}) enforces that $\pi_\Phi$ is a simple path that uses edges in $\mathcal{E}$, which can be achieved by introducing auxiliary flow conservation and degree constraints, commonly seen in linear program formulations for routing problems~\cite{miller1960integer}, rendering the above program nonlinear yet convex (see~\cite{marcucci2024shortest} for more details).
Constraints Eqn.~(\ref{eqn:gcs:in_set}) enforce that the two states $\mathbf{x}_v, \mathbf{y}_v$ regarding each vertex $v\in\mathcal{V}$ must reside within the corresponding collision-free convex set $\mathcal{X}_v$, ensuring that the trajectory segment $(\mathbf{x}_v, \mathbf{y}_v)$ is also collision-free.
Constraints Eqn.~(\ref{eqn:gcs:continuity}) enforce that the terminal state of $u$ always coincides with the initial state of $v$ for any edge $(u,v)$ of $\pi_\Phi$, ensuring that the reconstructed trajectory is continuous.
Constraints Eqn.~(\ref{eqn:gcs:init_cond}) ensure that $\tau$ starts from $\mathbf{x}_\text{start}$ and ends at $\mathbf{x}_\text{goal}$. Note that the two vertices (convex sets) $v_\text{start},v_\text{goal}\in\mathcal{V}$ are determined by iterating through all vertices $v\in\mathcal{V}$ to check whether $\mathbf{x}_\text{start}\in\mathcal{X}_{v}$ or $\mathbf{x}_\text{goal}\in\mathcal{X}_{v}$, respectively.\footnote{In case of multiple $v_\text{start}$ (or $v_\text{goal}$), a hyper vertex is created connecting itself to each $v_\text{start}$ (or $v_\text{goal}$)~\cite{bertsimas1997introduction} with slight changes to constraints Eqn.~(\ref{eqn:gcs:init_cond}).
}

The above formulation aligns with standard MRMP conventions, focusing on kinematic feasibility and omitting differential and kinodynamic constraints. If needed, these constraints can be incorporated by augmenting the state space and adding linear or other convex constraints to capture, for example, bounded accelerations or nonholonomic motion.


\subsection{ST-GCS: Time-Optimal Spatiotemporal Motion Planning}\label{subsec:gcs_time_ext}
Although GCS effectively handles single-robot motion planning in a time-invariant $d$-dimensional state space, spatiotemporal motion planning with dynamic obstacles requires additional machinery to handle dynamic avoidance, variable arrival times, velocity bounds, and time optimality in a unified optimization. We thus propose ST-GCS, which operates on a graph of collision-free space-time convex sets.

\begin{figure}
\centering
\includegraphics[width=\linewidth]{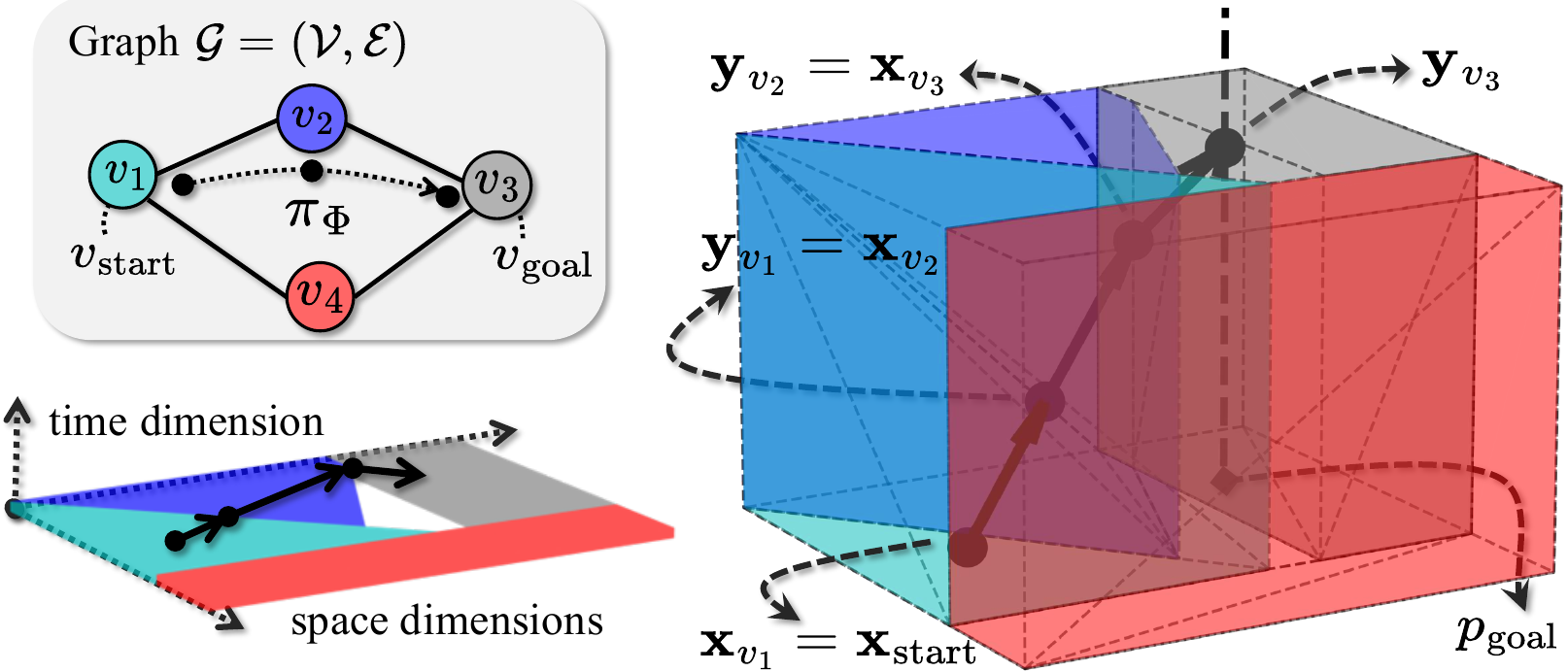}
\caption{Piecewise linear trajectories from $\mathbf{x}_\text{start}$ to $p_\text{goal}$ through collision-free 2D space convex sets (lower-left) and 3D space-time convex sets (right).}
\label{fig:tgcs}
\end{figure}

Let $\mathcal{G}=(\mathcal{V},\mathcal{E})$ represent a space-time decomposition, where each vertex $v\in\mathcal{V}$ corresponds to a space-time convex set $\mathcal{X}_v\subset\mathbb{R}^{d+1}$ that is free of both static and dynamic obstacles. 
We construct these sets by (1) extruding each given spatial convex set (free of static obstacles) from time $0$ to an arbitrarily large global time limit $t_{\text{max}}$ and (2) applying ECD (see Sec.~\ref{subsec:ecd}) to remove space-time regions intersecting other robots' trajectories (treated as dynamic obstacles), potentially subdividing the extruded sets further.
An edge $(u,v)\in\mathcal{E}$ indicates $\mathcal{X}_u\cap\mathcal{X}_v\neq\emptyset$.

Let $\mathbf{x}.p$ and $\mathbf{x}.t$ respectively denote the spatial and temporal components of a space-time state $\mathbf{x}$. Within each space-time convex set $\mathcal{X}_v$, the local trajectory segment $\mathbf{x}_v \rightarrow \mathbf{y}_v$ is traversed at a uniform velocity $\mathbf{v}=\dfrac{\mathbf{y}_v.p-\mathbf{x}_v.p}{\mathbf{y}_v.t-\mathbf{x}_v.t}$, potentially different across sets.
We now present the following ST-GCS formulation for spatiotemporal motion planning, which explicitly specifies a time-minimizing objective in Eqn.~(\ref{eqn:tgcs:obj}) and linear time and velocity constraints in Eqn.~(\ref{eqn:tgcs:forward_time}-\ref{eqn:tgcs:vel_bound}):
\begin{align}
\min_{\Phi,\mathbf{x},\mathbf{y}}\quad
& \sum_{v\in\pi_\Phi}(\mathbf{y}_v.t-\mathbf{x}_v.t)\label{eqn:tgcs:obj}\\
\textbf{s.t.}\quad
&\text{Constraints in Eqn.~(\ref{eqn:gcs:path}-\ref{eqn:gcs:continuity})}\\
&\mathbf{y}_v.t - \mathbf{x}_v.t > 0, &\forall v\in\pi_\Phi\label{eqn:tgcs:forward_time}\\
&\mathbf{v}_\text{min} \preceq \dfrac{\mathbf{y}_v.p-\mathbf{x}_v.p}{\mathbf{y}_v.t - \mathbf{x}_v.t}\preceq \mathbf{v}_\text{max},&\forall v\in\pi_\Phi\label{eqn:tgcs:vel_bound}\\
&\mathbf{x}_{v_\text{start}}=\mathbf{x}_\text{start}, \mathbf{y}_{v}.p=p_\text{goal}&\text{with }v\in\mathcal{V}_\text{goal}\label{eqn:tgcs:init_cond}.
\end{align}
where Constraints Eqn.~(\ref{eqn:tgcs:forward_time}) prevents time reversal from each state to a subsequent state
\footnote{In practice, we impose Eqn.~(\ref{eqn:tgcs:forward_time}) as $\mathbf{y}_v.t - \mathbf{x}_v.t \geq \epsilon$ with an inclusive lower bound of a small positive number $\epsilon$.}.
Constraints Eqn.~(\ref{eqn:tgcs:vel_bound}) impose given velocity bounds in each spatial dimension.
Unlike Eqn.~(\ref{eqn:gcs:init_cond}) in the static GCS formulation, Constraint Eqn.~(\ref{eqn:tgcs:init_cond}) only enforces that the spatial component matches the given goal position $p_\text{goal}$, leaving the arrival time unconstrained. 
Note that, while $v_\text{start}$ is determined in the same way as in GCS, each goal vertex $v\in\mathcal{V}_\text{goal}$ can be identified by checking whether $\mathcal{X}_{v}$ contains any states with the goal position (i.e., $\mathcal{X}_{v} \cap \{(p_\text{goal},t)\,|\,0\leq t\leq t_\text{max}\}\neq\emptyset$) and, if so, whether any arrival time at the goal position within $\mathcal{X}_{v}$ can be extended to $t_\text{max}$ (i.e., $\{(p_\text{goal},t)\,|\,t^*\leq t\leq t_\text{max}\}\subseteq \bigcup_{u\in\mathcal{V}}\mathcal{X}_u$, where $t^*=\min\{t\mid (p_\text{goal},t)\in \mathcal{X}_v\}$) to ensure the robot can stay there indefinitely.

As shown in Fig.~\ref{fig:tgcs}, given a feasible solution that fixes the values of $\pi_\Phi$, $\mathbf{x}$, and $\mathbf{y}$, we can reconstruct a space-time piecewise linear trajectory, and any space-time state along the dashed line can serve as a valid goal if the robot can indefinitely stay at $p_\text{goal}$ the arrival. 
In summary, ST-GCS fuses the core convex path parametrization of GCS with explicit time and velocity constraints, enabling time-optimal
\footnote{More rigorously, the time optimality is defined over the solution space of all piecewise linear trajectories parameterized by $(\Phi,\mathbf{x},\mathbf{y})$.}, piecewise-linear trajectories in a space-time domain that can include dynamic obstacles (handled via ECD).

 \section{ST-GCS For Multi-Robot Motion Planning}
In this section, we consider MRMP with $n$ robots, where each robot $i$ is given a start state $\mathbf{x}_\text{start}^{(i)}$ and a goal position $p_\text{goal}^{(i)}$. 
The problem is to compute $n$ space-time collision-free trajectories, $\mathcal{T}=\{\tau_i\}_{i=1}^n$, for the robots.
Our approach relies on the Exact Convex Decomposition (ECD) algorithm (Sec.~\ref{subsec:ecd}), which ``reserves'' a given piecewise linear trajectory on a graph $\mathcal{G}$ of space-time convex sets, thereby producing an updated graph $\mathcal{G}'$ whose convex sets are collision-free with respect to the reserved trajectory. Consequently, once some robots' trajectories are planned, they can be treated as dynamic obstacles for the remaining robots by applying ECD and then running ST-GCS on $\mathcal{G}'$. 
Following standard MAPF practices, we solve MRMP via two prioritized planning frameworks (Sec.~\ref{subsec:alg}), i.e., Random-Prioritized Planning (RP) and Priority-Based Search (PBS), where robots plan one at a time while avoiding collisions with higher-priority robots’ trajectories, which are incorporated as dynamic obstacles through ECD. 

\begin{figure*}
\centering
\includegraphics[width=0.92\linewidth]{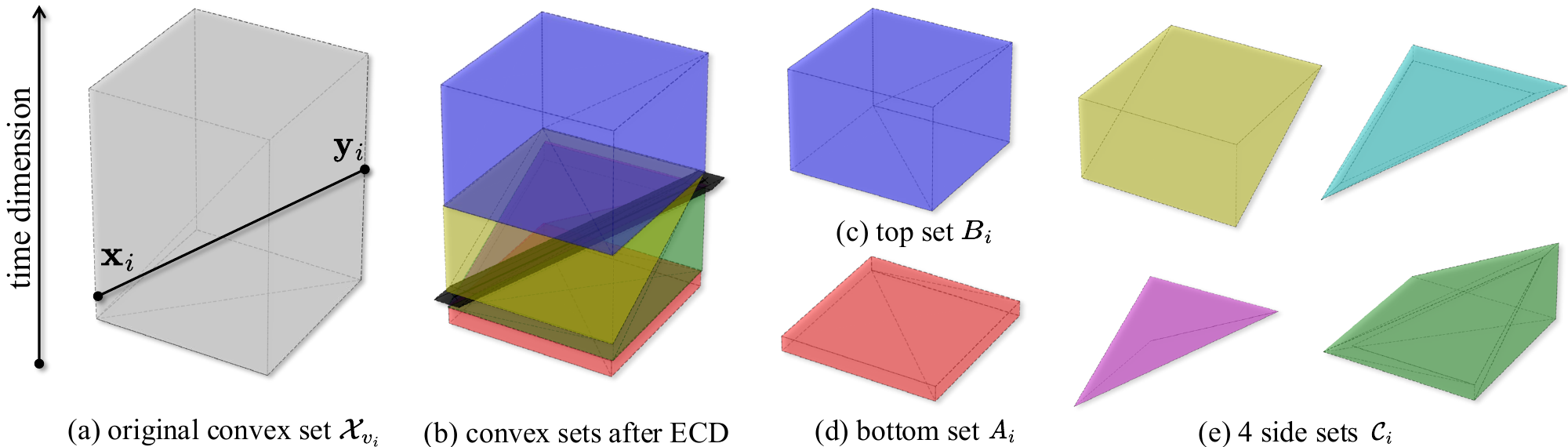}
\caption{The proposed ECD algorithm decomposes (a) convex set $\mathcal{X}_{v_i}$ for a vertex $v_i\in\mathcal{V}$ and a trajectory segment $\mathbf{x}_i \rightarrow \mathbf{y}_i$ into (c-e) $6$ convex sets. (b) The trajectory segment is enlarged to a parallelepiped (highlighted in black) with its apothem being the safe radius between robots.}
\label{fig:ecd}
\end{figure*}

\subsection{Exact Convex Decomposition (ECD)}\label{subsec:ecd}
ECD takes as input a sequence of vertex-segment tuples $\{(v_i,\mathbf{x}_i,\mathbf{y}_i)\}_{i=1}^k$, where each trajectory segment $\mathbf{x}_i \rightarrow \mathbf{y}_i$ lies entirely in a single convex set $\mathcal{X}_{v_i}$ of the given graph $\mathcal{G}$.
We will discuss how to obtain such a sequence from any given piecewise linear trajectory resulting from an ST-GCS solution at the end of this subsection. 
Here, we assume each robot has a $d$-dimensional hypercube occupancy centered at each state $\mathbf{x}$ and exemplify ECD in a 3D space-time state space, but the same principle can be extended to other convex-shaped occupancies and to higher-dimensional state spaces with an additional time dimension.

\noindent\textbf{Pseudocode (Alg.~\ref{alg:ecd}):}
ECD iterates over the $k$ vertex-segment tuples [Line~\ref{alg:ecd:iter_start}].
To simplify notation, let $\boxdot(p,r)$ denote the 2D square centered at $p$ with apothem $r$.
For each tuple $(v_i,\mathbf{x}_i,\mathbf{y}_i)$, ECD first constructs a \textit{parallelepiped} $L_i$ with its top and bottom faces being squares $\boxdot\left(\mathbf{x}_i.p,r\right)$ and $\boxdot\left(\mathbf{y}_i.p,r\right)$, respectively (see Fig.~\ref{fig:ecd}(b)).
ECD then creates two convex sets $A_i$ and $B_i$ [Lines~\ref{alg:ecd:Ai}-\ref{alg:ecd:Bi}], bounding $L_i$ from above and below (see Fig.~\ref{fig:ecd}(b)).
Specifically, $B_0$ excludes the cuboid formed by extruding $\boxdot\left(\mathbf{x}_0.p,r\right)$ from time $0$ to $\mathbf{x}_0.t$, and $A_k$ excludes the cuboid formed by extruding $\boxdot\left(\mathbf{x}_k.p,r\right)$ from time $\mathbf{x}_k.t$ to $t_\text{max}$. 
This is crucial as it ensures the robot can safely stay at $\mathbf{x}_0.p$ during $t\in[0,\mathbf{x}_0.t)$ and at $\mathbf{x}_k.p$ during $t\in[\mathbf{x}_k.t,t_\text{max})$.
As shown in Fig.~\ref{fig:ecd}(e), after removing $A_i\cup B_i$ from $\mathcal{X}_{v_i}$, ECD partitions the remaining center region into four convex sets by intersecting it with each halfspace $\text{outside}(L_i,s)$ [Line~\ref{alg:ecd:slice_side}], where $s$ is each side face of $L_i$, considered as a separating plane, and $\text{outside}(L_i,s)$ is the halfspace that does not contain $L_i$ (i.e., $\text{outside}(L_i,s)\cap L_i=\emptyset$). ECD then removes $\text{outside}(L_i,s)$ from the remaining region before processing the next halfspace [Line~\ref{alg:ecd:exclude_outside}]. 
After partitioning $\mathcal{X}_{v_i}$ into these new convex sets, ECD adds a vertex corresponding to each new set to $\mathcal{G}$ [Line~\ref{alg:ecd:add_vert}].
After processing all $k$ tuples, ECD updates the edges of $\mathcal{G}$ by checking intersections among new neighboring sets [Line~\ref{alg:ecd:update_edges}].
Fig.~\ref{fig:ecd} shows an example for an intermediate tuple $(v_i,\mathbf{x}_i,\mathbf{y}_i)$ with $1<i<k$.

\begin{algorithm}[t]
\DontPrintSemicolon
\linespread{0.95}\selectfont
\caption{ECD for Trajectory Reservation}\label{alg:ecd}
\SetKwInput{KwInput}{Input}
\KwInput{vertex-segment tuples $\{(v_i,\mathbf{x}_i,\mathbf{y}_i)\}_{i=1}^k$, graph $\mathcal{G}=(\mathcal{V},\mathcal{E})$ of convex sets, safe radius $r$}
\For{$i=1,2,...,k$}{\label{alg:ecd:iter_start}
    $A_{i}\gets\{\mathbf{x}\in\mathcal{X}_{v_i}\,|\,\mathbf{x}.t\leq \mathbf{x}_{i}.t\}$\Comment{Fig.~\ref{fig:ecd}(d)}\;\label{alg:ecd:Ai}
    $B_{i}\gets\{\mathbf{x}\in\mathcal{X}_{v_i}\,|\,\mathbf{x}.t\geq \mathbf{y}_i.t\}$\Comment{Fig.~\ref{fig:ecd}(c)}\;\label{alg:ecd:Bi}
    $\mathcal{C}_{i}\gets\{\}$,\, $\mathcal{X}_{v_i}\gets\mathcal{X}_{v_i}\setminus (A_{i}\cup B_{i})$\;
    $L_i\gets$ the parallelepiped defined by $\mathbf{x}_i,\mathbf{y}_i$, $r$\;
    \For{$s\in$ $\{\text{side faces of parallelepiped } L_i \}$}{
        $\mathcal{C}_{i}\gets\mathcal{C}_{i}\cup\{\mathcal{X}_{v_i}\cap \text{outside}(L_i,s)\}$\Comment{Fig.~\ref{fig:ecd}(e)}\;\label{alg:ecd:slice_side}
        $\mathcal{X}_{v_i}\gets\mathcal{X}_{v_i}\setminus \text{outside}(L_i,s)$\;\label{alg:ecd:exclude_outside}
    }
    add a vertex for every set in $\mathcal{C}_{i}\cup\{A_{i},B_{i}\}$ to $\mathcal{G}$\;\label{alg:ecd:add_vert}
}
update edges in $\mathcal{G}$ by checking set intersections\;\label{alg:ecd:update_edges}
\end{algorithm}

\noindent\textbf{Vertex-Segment Sequence Construction:}
We describe how to construct a sequence $S=\{(v_i,\mathbf{x}_i,\mathbf{y}_i)\}_{i=1}^k$ of vertex-segment tuples from a piecewise linear trajectory $\tau$ on $\mathcal{G}$.
This construction is particularly relevant in our prioritized planning frameworks (Sec.~\ref{subsec:alg}) and for dynamic obstacles, where ECD reserves trajectories for collision avoidance. 
Three cases arise: 
(1) If $\tau$ is reconstructed from an ST-GCS solution computed on the same graph $\mathcal{G}$, each segment of $\tau$ naturally maps to a unique convex set $\mathcal{X}_v$ of $\mathcal{G}$, and potentially to the adjacent convex sets of $\mathcal{X}_v$ if the segment lies in the boundaries of $\mathcal{X}_v$.
(2) If a segment of $\tau$ intersects more than one convex set of $\mathcal{G}$, we subdivide it at each boundary so that each sub-segment lies in exactly one convex set. 
(3) If a convex set $\mathcal{X}_{v}$ contains multiple segments $\mathbf{x}_j \rightarrow \mathbf{y}_j$ (non-overlapping in time), we slice $\mathcal{X}_{v}$ by the planes $t=\mathbf{x}_j.t$ and $t=\mathbf{y}_j.t$, ensuring each resulting subset contains at most one segment.
We adopt a unified procedure to construct the sequence $S$, considering the above three cases.
For each linear segment in $\tau$, we check whether it intersects with each convex set of $\mathcal{G}$\footnote{
Our implementation leverages bounding boxes for the convex sets to do coarse collision checking before exact collision checking for efficiency.
}.
If intersecting, we collect the vertex $v_i$ and the two endpoints $\mathbf{x}_i,\mathbf{y}_i$ of the intersecting sub-segment into sequence $S$.

\subsection{Prioritized Planning Frameworks}~\label{subsec:alg}
We now introduce two prioritized planning frameworks for MRMP that use ST-GCS as the single-robot planner and ECD as a subroutine for trajectory reservation.

\begin{algorithm}[t]
\DontPrintSemicolon
\linespread{0.95}\selectfont
\caption{RP+ST-GCS for MRMP}\label{alg:rp}
\SetKwInput{KwInput}{Input}
\KwInput{start and goal pairs$\{(\mathbf{x}_\text{start}^{(i)},p_\text{goal}^{(i)} )\}_{i=1}^n$,\quad\quad graph $\mathcal{G}$ of convex sets}
\While{not reaching the terminal condition}{
    $\mathcal{G}'\gets$ a copy of $\mathcal{G}$\;\label{alg:RP:copy_G}
    \For{$i\in \text{random unused permutation of } \{i\}_{i=1}^n$}{\label{alg:RP:pp}
        $\tau_i\gets$ \text{solve}($\mathcal{G}',\mathbf{x}_\text{start}^{(i)},p_\text{goal}^{(i)})$\;\label{alg:RP:replan}
                \If{solving reports failure}{
                    \textbf{break}\;
                }
        $\mathcal{G}'\gets$ reserve $\tau_i$ on $\mathcal{G}'$ via ECD\;\label{alg:RP:ecd}
    }
    \If{ST-GCS solving succeed for all robots}{
        \Return $\{\tau_i\}_{i=1}^n$\;\label{alg:RP:return}
    }
}
\Return ``\textit{fail to find a solution}''\;\label{alg:RP:fail}
\end{algorithm}

\noindent\textbf{Random-Prioritized Planning (RP):}
RP (Alg.~\ref{alg:rp}) explores random total priority orders (i.e., permutations of the robots). For each order, the robots plan sequentially [Line~\ref{alg:RP:pp}]. 
Each robot $i$ plans its trajectory $\tau_i$ by solving ST-GCS on the current graph $\mathcal{G}'$ [Line~\ref{alg:RP:replan}], which is then updated with the planned $\tau_i$ reserved via ECD [Line~\ref{alg:RP:ecd}]. 
RP returns the first feasible solution once all robots successfully plan their trajectories [Line~\ref{alg:RP:return}]. Otherwise, it attempts the next order, until a terminal condition (e.g., runtime limit) is met.

\noindent\textbf{Priority-Based Search (PBS):} 
PBS (Alg.~\ref{alg:pbs}) systematically explores priority orders to resolve collisions by searching a priority tree, where each node $N$ contains a unique priority set $\boldsymbol{\pmb\prec}_N$ of ordered pairs of robots and a set $N.\mathcal{T}$ of $n$ trajectories that respect the prioritized planning scheme specified by $\boldsymbol{\pmb\prec}_N$. 
PBS initializes the root node with an empty priority set and potentially colliding trajectories [Lines~\ref{alg:PBS:init_root}-\ref{alg:PBS:root_plan}].
When expanding a node $N$, PBS
checks $N.\mathcal{T}$ for collisions [Line~\ref{alg:PBS:col_check}].
If none are found, then it returns $N.\mathcal{T}$ as a solution [Line~\ref{alg:PBS:sol_found}]. 
Otherwise, it identifies a colliding pair $\tau_i$ and $\tau_j$ [Line~\ref{alg:PBS:1st_col}] and generates two child nodes $N_1$ and $N_2$, adding the pair $i\prec j$ ($i$ has a higher priority than $j$) to $\boldsymbol{\pmb\prec}_{N_1}$ and $j\prec i$ to $\boldsymbol{\pmb\prec}_{N_2}$ [Lines~\ref{alg:PBS:for_child}-\ref{alg:PBS:update_prec}].
For each child node $N'$, PBS invokes \texttt{UpdateNode} to replan the trajectories for a list of robots [Line~\ref{alg:PBS:replan_list}] to ensure that all trajectories in $N'.\mathcal{T}$ respect $\boldsymbol{\pmb\prec}_{N'}$.
With all the high-priority trajectories reserved on $\mathcal{G}$ via ECD [Line~\ref{alg:PBS:ecd}], each replanning calls ST-GCS for a lower-priority robot $j$ [Line~\ref{alg:PBS:replan}].
If \texttt{UpdateNode} succeeds, PBS pushes the child node $N'$ to the stack top [Lines~\ref{alg:PBS:update_node}-\ref{alg:PBS:add_child}].
PBS returns failure if no valid solution can be found after visiting all possible nodes in the priority tree [Line~\ref{alg:PBS:fail}].

\noindent\textbf{ST-GCS Solving:}
Solving ST-GCS to optimality can be expensive, especially after many ECD updates that enlarge the graph. Therefore, as a low-level planner for RP and PBS, we employ the convex-restriction and path-restriction heuristic approach from~\cite{marcucci2023motion} rather than seeking global optimality~\cite{marcucci2024shortest}. 
In short, this approach first relaxes binary edge variables to fractional values, then heuristically reconstructs a graph path $\pi_\Phi$ by interpreting each fractional $\phi_e$ as the probability of using edge $e$. 
Fixing $\pi_\Phi$ yields a final convex program for ST-GCS that can be solved at a relatively low computational cost. 
Re-running this procedure multiple times with different random seeds can further improve solution quality, returning the best trajectory found.

\begin{algorithm}[t]
\DontPrintSemicolon
\linespread{0.95}\selectfont
\caption{PBS+ST-GCS for MRMP}\label{alg:pbs}
\SetKwInput{KwInput}{Input}
\KwInput{start and goal pairs$\{(\mathbf{x}_\text{start}^{(i)},p_\text{goal}^{(i)} )\}_{i=1}^n$, \quad\quad graph $\mathcal{G}$ of convex sets}
create a root node $N_\text{root}$ w/ $\boldsymbol{\pmb\prec}_{N_\text{root}}\gets\emptyset$\;\label{alg:PBS:init_root}
$N_\text{root}.\mathcal{T}\gets\{\text{solve}(\mathcal{G},\mathbf{x}_\text{start}^{(i)},p_\text{goal}^{(i)})\}_{i=1}^n$\;\label{alg:PBS:root_plan}
Stack $\gets\{N_\text{root}\}$\;
\While{Stack $\neq\emptyset$}{
    $N\gets$ Stack.pop()\;
    \If{no collisions in $N.\mathcal{T}$}{\label{alg:PBS:col_check}
        \Return $N.\mathcal{T}$\;\label{alg:PBS:sol_found}
    }
    $\tau_i,\tau_j\gets$ first pairwise collision in $N.\mathcal{T}$\;\label{alg:PBS:1st_col}
    \For{$(i,j)\in\{(i,j),(j,i)\}$}{\label{alg:PBS:for_child}
        create node $N'$ w/ $\boldsymbol{\pmb\prec}_{N'}\gets\boldsymbol{\pmb\prec}_{N}\cup\{i\prec j\}$\;\label{alg:PBS:update_prec}
        \If{\texttt{UpdateNode}$(N', \mathcal{G},j)$}{\label{alg:PBS:update_node}
             Stack.push($N'$)\;\label{alg:PBS:add_child}
        }
    }
}
\Return ``\textit{fail to find a solution}''\;\label{alg:PBS:fail}
\SetKwFunction{FMain}{UpdateNode}
\SetKwProg{Fn}{Function}{:}{}
\Fn{\FMain{$N, \mathcal{G}, i$}}{
    $L\gets\{i\}\cup\{j\,|\,1\leq j\leq n, i\prec_N j\}$\;\label{alg:PBS:replan_list}
    \For{$j\in topologicalSort(L,\boldsymbol{\pmb\prec}_N)$}{
        \If{$\exists k\prec_N j \textbf{ s.t. }N.\tau_k \text{ collides w/ } N.\tau_j$}{
            $\mathcal{G}'\gets$ reserve $\{\tau_k|k\prec_N j\}$ on $\mathcal{G}$ via ECD\;\label{alg:PBS:ecd}
            $N'.\tau_{j}\gets$ \text{solve}($\mathcal{G}',\mathbf{x}_\text{start}^{(j)},p_\text{goal}^{(j)})$\;\label{alg:PBS:replan}
            \If{solving reports failure}{
                \Return \textbf{False}\Comment{fails to update $N$}\;
            }
        }
    }
\Return \textbf{True}\Comment{succeeds to update $N$}\;
}
\end{algorithm}

\section{Experiments}
This section presents our experimental results on an \textit{Apple}\textsuperscript{\textregistered} M4 CPU machine with 16GB RAM. 
We evaluate \textbf{RP+ST-GCS} and \textbf{PBS+ST-GCS} against three baseline methods in 2D mobile robot domains.
All methods are implemented in Python, and our ST-GCS program is solved using the \textit{Drake}~\cite{drake} library with the Mosek\footnote{\url{https://www.mosek.com/}} solver.
The source code and numerical results are publicly available on \url{https://github.com/reso1/stgcs}.
More detailed visualizations and simulation videos of our approach can be found at \url{https://sites.google.com/view/stgcs}.

\subsection{Experiment Setup}

\begin{figure*}
\centering
\includegraphics[width=\linewidth]{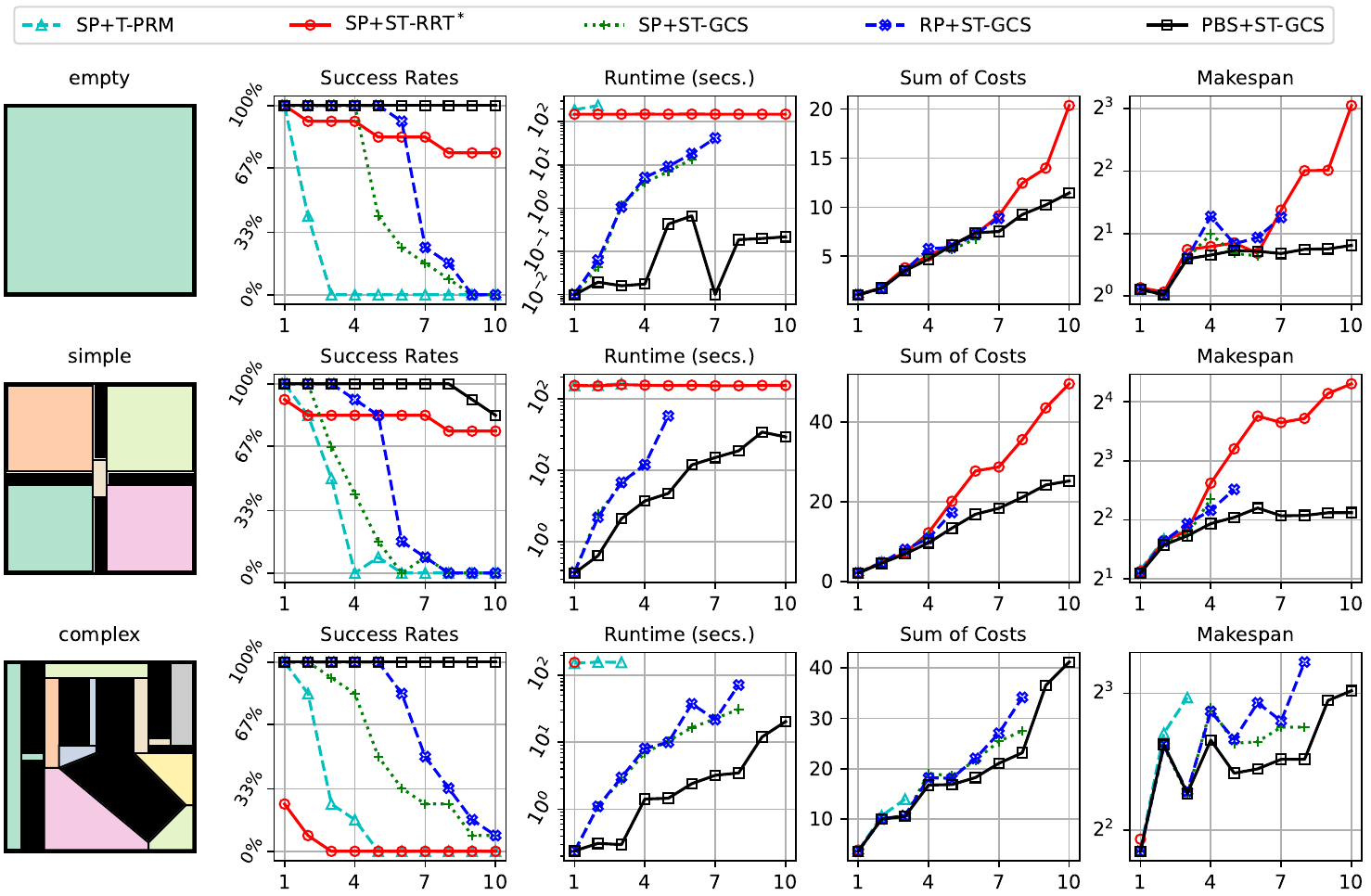}
\caption{Results for all MRMP methods on three maps. Each subplot shows the number of robots (x-axis) versus a specific metric (y-axis).}
\label{fig:stats}
\end{figure*}

\noindent\textbf{Instances:}
We use three benchmark 2D maps (Fig.~\ref{fig:stats}): \textit{empty}, \textit{simple}, and \textit{complex}~\cite{marcucci2023motion}. 
Their spatial collision-free convex sets are given; we extrude each along time $t\in [0,50]$ to create space-time convex sets. 
For \textit{simple}, we additionally introduce four dynamic disk obstacles moving at constant velocities, then apply ECD to reserve their trajectories.
For each map and each $n \in [1,10]$ of robots, we generate 12 random instances by sampling start states and goal positions within the space-time convex sets.

\noindent\textbf{Baselines:}
We compare against three sequential planning (SP) baselines---i.e., prioritized planning with the fixed priority order by robot index---each with a different low-level single-robot planner.
(1) (Adapted) \textbf{SP+T-PRM}~\cite{huppi2022t}: We adapt T-PRM to support collision checks on roadmap edges when finding shortest paths (missed by the original implementation) and restrict its random sampling to the given spatial convex sets.
(2) (Adapted) \textbf{SP+ST-RRT$^*$}~\cite{grothe2022st}: We restrict the random sampling of ST-RRT$^*$ random sampling to the given spatial convex sets.
(3) \textbf{SP+ST-GCS}.

\noindent\textbf{Parameters:} 
For \textit{empty}, we set a maximum velocity limit of $0.5$ for each space dimension; For \textit{empty} and \textit{complex}, we set a maximum velocity limit of $1.0$ for each spatial dimension.
We set a runtime limit of 150 seconds for all methods. SP+T-PRM allocates $150/n$ seconds for each robot's PRM construction (omitting its shortest-path finding runtime), and SP+ST-RRT$^*$ allocates $150/n$ seconds for each single-robot ST-RRT$^*$ planning.
For ST-GCS solving on any graph $\mathcal{G}=(\mathcal{V},\mathcal{E})$ using the heuristic approach (see Sec.~\ref{subsec:alg}), we limit the number of graph paths sampled to $1e^3\times\log|\mathcal{E}|$.

\subsection{Results and Analysis}

Fig.~\ref{fig:stats} reports four metrics: \textbf{Success Rate} (out of 12 random instances),  \textbf{Runtime} (in seconds, on a \textbf{log-scaled} y-axis), \textbf{SoC} (the sum of time costs of all trajectories), and \textbf{Makespan} (the maximum time costs of all trajectories, on a \textbf{log2-scaled} y-axis).
The last three metrics are averaged only over the intersection of instances solved by each method, ensuring a fair comparison. If a method solves fewer than 3 of 12 instances for a given $n$, it is excluded from that average to avoid empty intersections.

\noindent\textbf{Success Rates and Runtimes:}
PBS+ST-GCS solves all instances on \textit{empty} and \textit{complex}, with average runtimes consistently under 1 second and 10 seconds on average, respectively---often orders of magnitude faster than sampling-based methods. Adding dynamic obstacles on \textit{simple} significantly enlarges the initial space-time graph (via ECD), which can slow ST-GCS solving; nevertheless, PBS+ST-GCS still achieves the highest success rates overall.
RP+ST-GCS randomly explores different priority orders until timeout generally attains higher success rates than SP+ST-GCS which uses only a fixed priority order.
Among sampling-based methods, SP+ST-RRT$^*$ outperforms SP+T-PRM on \textit{empty} and \textit{simple}, likely due to the inherent faster tree-based exploration rooted at the starts and goals. However, on \textit{complex} with many spatial corridors, the more global random exploration of T-PRM yields higher success rates than ST-RRT$^*$.
Still, both fail when $n>2$ on \textit{complex}.
Comparing all SP-based methods, SP+ST-GCS typically runs faster than the two sampling-based methods and achieves higher success rates in settings with more spatial corridors. However, its success rates are lower for larger $n$, due to the heuristic solver failing more frequently, as analyzed in the ablation study below. 

\noindent\textbf{Solution Quality:}
PBS+ST-GCS consistently yields the best solution quality across all maps. Its SoC scales almost linearly with $n$, and its makespan barely increases as $n$ grows. Other ST-GCS variants also produce better solutions than the sampling-based methods in many cases.

\begin{figure}
\centering
\includegraphics[width=\linewidth]{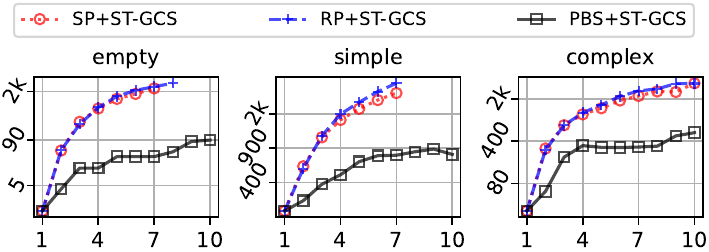}
\caption{Average number of graph edges in SP+ST-GCS, RP+ST-GCS, and PBS+ST-GCS when each returns a feasible solution.}
\label{fig:graph_size}
\end{figure}

\begin{figure}
\centering
\includegraphics[width=\linewidth]{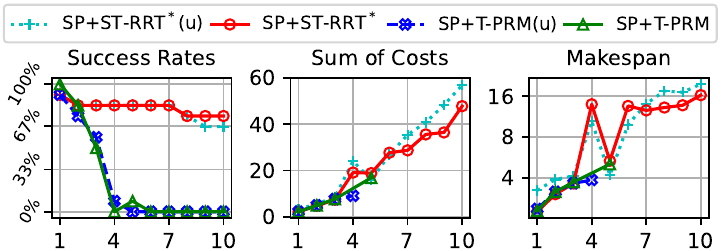}
\caption{Comparing the two sampling-based approaches with their sampling procedures performed on map bounding box and space collision-free sets.}
\label{fig:sampling}
\end{figure}

\noindent\textbf{Ablation---Effectiveness of PBS for ST-GCS:}
Our raw data indicates that large space-time graphs (over 1000 edges) resulting from numerous dynamic obstacles can impede the heuristic solver under limited path-sampling budgets. Fig.~\ref{fig:graph_size} demonstrates that PBS mitigates this issue by reserving only conflicting trajectories deemed high-priority, controlling the graph size growth more effectively than SP+ST-GCS and RP+ST-GCS (which both rely on total priority orders with sequential ECD, thus expanding the graph faster as $n$ grows).

\noindent\textbf{Ablation---Constrained Sampling:}
Fig.~\ref{fig:sampling} compares T-PRM and ST-RRT$^*$ with random sampling constrained to the spatial collision-free convex sets versus unconstrained sampling (labeled \textbf{T-PRM(u)} and \textbf{ST-RRT$^*$(u)}) over the entire bounding box. Constrained sampling yields modest improvements in success rates (though T-PRM still fails for higher $n$) and better solution quality for ST-RRT$^*$. This confirms that focusing sampling on collision-free regions can be beneficial, albeit insufficient to match ST-GCS performance.

\subsection{Case Study}
\begin{figure}
\centering
\includegraphics[width=0.98\linewidth]{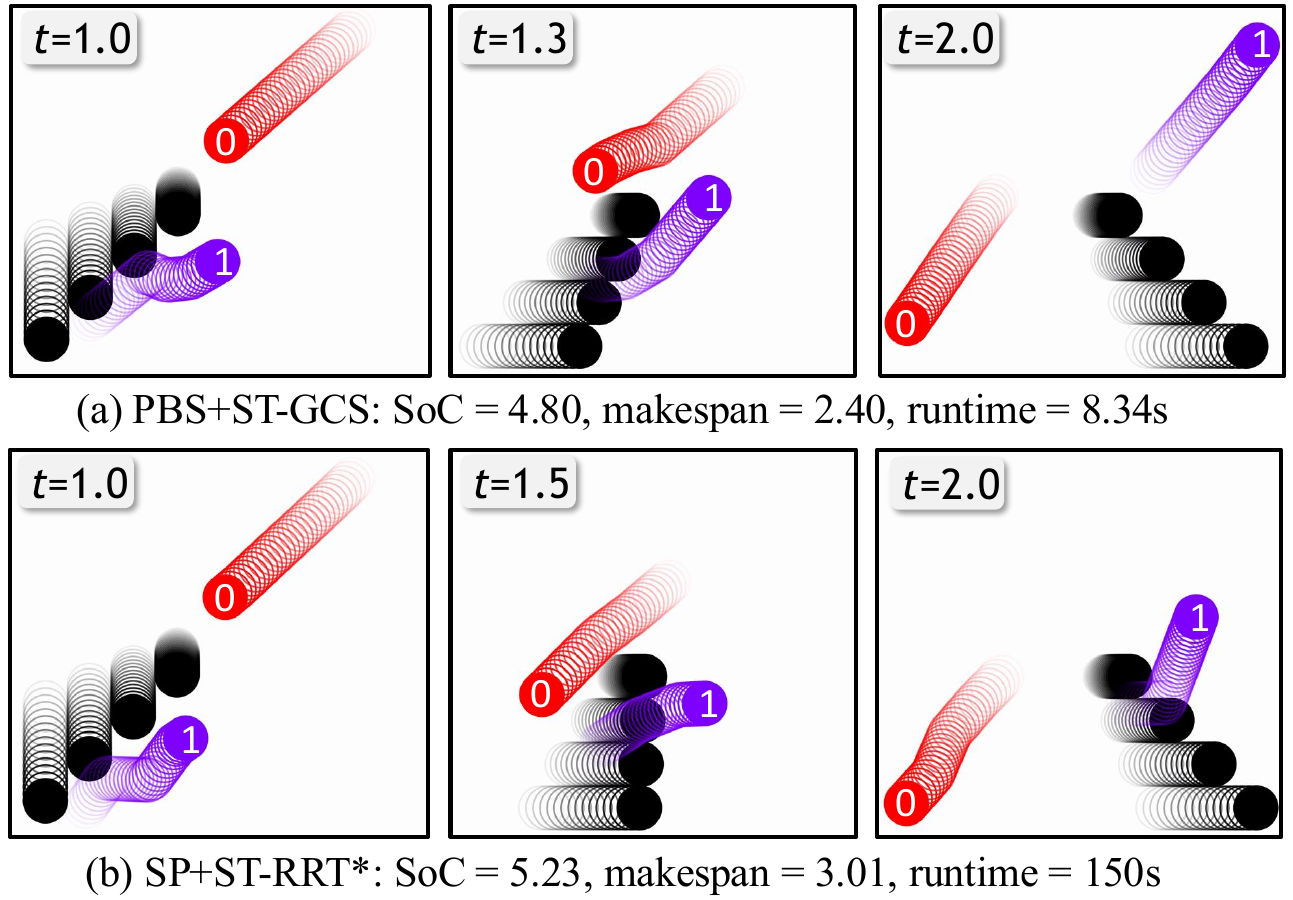}
\caption{Two robots exchange positions with four dynamic obstacles (black).}
\label{fig:demo_dynamic_obstacles}
\end{figure}

\begin{figure}
\centering
\includegraphics[width=\linewidth]{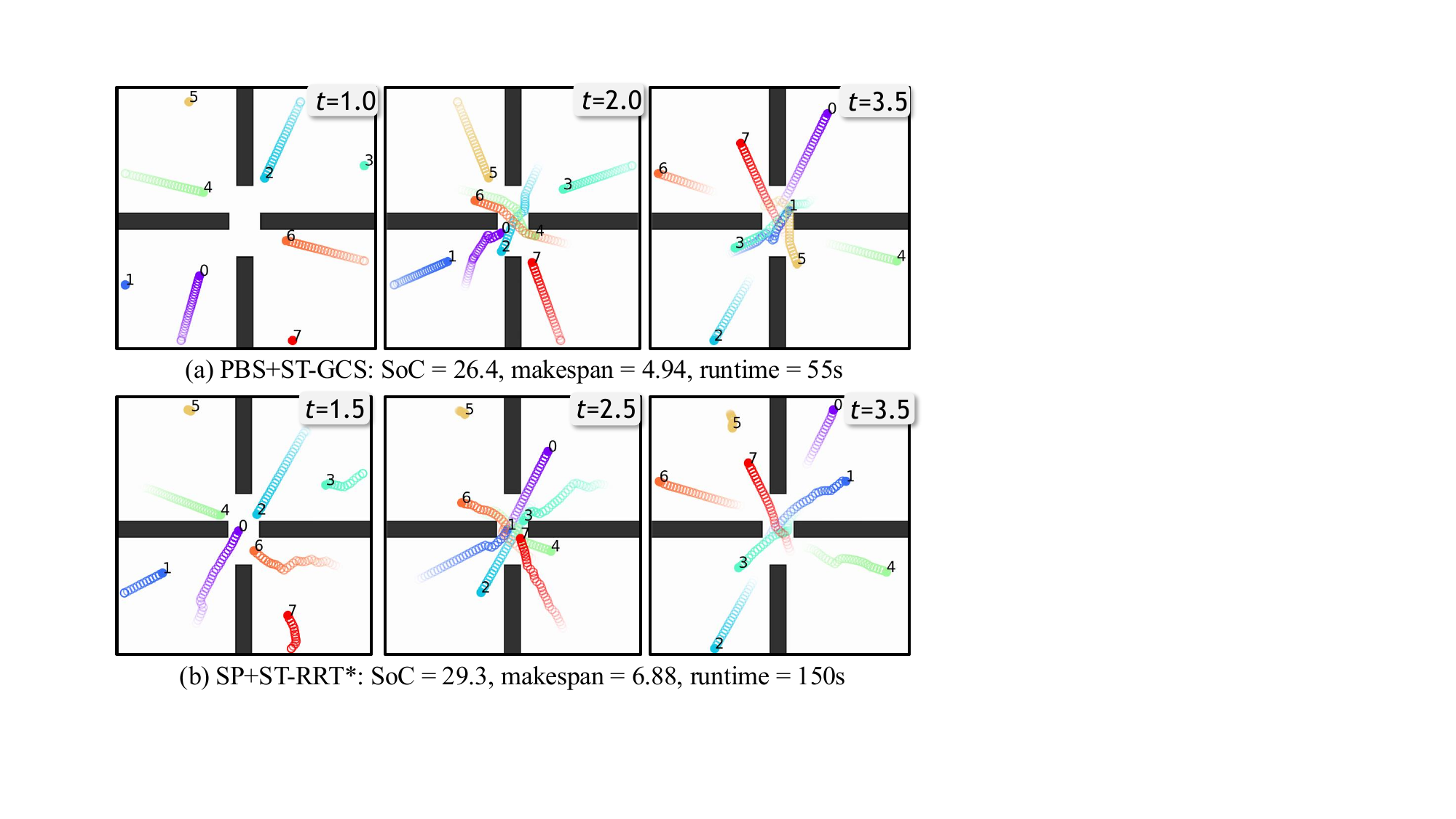}
\caption{Eight robots exchange positions with one another on \textit{simple}.}
\label{fig:demo_mrmp}
\end{figure}

We highlight two examples demonstrating the advantages of PBS+ST-GCS over SP+ST-RRT$^*$.
Fig.~\ref{fig:demo_dynamic_obstacles} demonstrates that PBS+ST-GCS identifies a safe shortcut through dynamic obstacles, producing a time-optimal MRMP solution.
Fig.~\ref{fig:demo_mrmp} shows that PBS+ST-GCS enables robots to traverse the congested central region simultaneously with minimal waiting or detours, while SP+ST-RRT$^*$ generates zigzagging trajectories and significant delays (e.g., robot 5 is delayed until other robots nearly reach their goals).

\section{Conclusions \& Future Work}
We presented ST-GCS, a time-optimal deterministic approach that addresses the inherent limitations of sampling-based methods in spatiotemporal settings. 
By extending the GCS formulation to a space-time domain, ST-GCS systematically covers spatiotemporal bottlenecks using collision-free convex sets. 
Our ECD algorithm further enables straightforward reservation of piecewise linear trajectories of dynamic obstacles for collision avoidance. 
We integrated ST-GCS and ECD into prioritized planning frameworks for MRMP.
We demonstrated through extensive experiments that our approach consistently outperforms state-of-the-art sampling-based methods in both success rates and solution quality, often with orders-of-magnitude faster runtimes, especially in challenging scenarios such as narrow corridors and crowded environments.
Future work includes investigating robust pruning strategies for managing the ST-GCS graph size when numerous trajectories are reserved via ECD, generalizing to high-dimensional nonlinear configuration spaces and more complex obstacle models, and extending ST-GCS with richer motion models with nonholonomic and dynamic constraints.

\section*{Acknowledgement}
This work was supported by the NSERC under grant number RGPIN2020-06540 and a CFI JELF award. We thank the anonymous reviewers for their constructive feedback that helped us improve this paper.

\bibliographystyle{IEEEtran}
\bibliography{ref}

\end{document}